\documentclass[letterpaper, 10 pt, conference]{ieeeconf}  
\IEEEoverridecommandlockouts                              
\usepackage{graphicx}
\usepackage{bm}
\usepackage{cite}
\usepackage{multirow}
\usepackage{threeparttable}
\usepackage{hyperref}

\newcommand{\etal}{\textit{et al}. }
\usepackage{color}

\newcommand{\ourmethod}[0]{{\textit{VFAS-Grasp}}}
\title{\LARGE \bf
VFAS-Grasp: Closed Loop Grasping with Visual Feedback and Adaptive Sampling
}

\author{Pedro Piacenza*, Jiacheng Yuan*, Jinwook Huh and Volkan Isler
\thanks{All authors are with the Samsung AI Center NY, 30 W 26th St, New York, NY 10010. (e-mail: \texttt{$\{$p.piacenza, jinwook.huh, ibrahim.i$\}$@samsung.com, j.yuan@partner.samsung.com}).}%
\thanks{* These authors contributed equally.}%
}

\begin{document}
\maketitle
\thispagestyle{empty}
\pagestyle{empty}
\setlength{\parskip}{0pt}

\begin{abstract}
We consider the problem of closed-loop robotic grasping and present a novel planner which uses Visual Feedback and an uncertainty-aware Adaptive Sampling strategy (VFAS) to close the loop. At each iteration, our method \ourmethod{} builds a set of candidate grasps  by generating random perturbations of a seed grasp. The candidates are then scored using a novel metric which combines a learned grasp-quality estimator, the uncertainty in the estimate and the distance from the seed proposal to promote temporal consistency. Additionally, we present two mechanisms to improve the efficiency of our sampling strategy: We dynamically scale the sampling region size and number of samples in it based on past grasp scores. We also leverage a motion vector field estimator to shift the center of our sampling region. We demonstrate that our algorithm can run in real time (20 Hz) and is capable of improving grasp performance for static scenes by refining the initial grasp proposal. We also show that it can enable grasping of slow moving objects, such as those encountered during human to robot handover. Video: \url{https://youtu.be/8DRe2OFlf7o}
\end{abstract}


\section{Introduction}
A traditional robotic grasping pipeline typically uses an external camera which provides a scene representation as input to a grasp planner which then proposes a set of candidate grasps either for the full scene or for a target object. More often than not, the execution of one of these grasps is carried out in an open loop fashion, with little or no new sensor information used after the selection of the best grasp candidate. Under such circumstances, grasping may fail due to poor pose or object shape estimation, camera calibration and other perception artifacts.

A closed loop control system periodically incorporates sensor data as a task progresses, computes an error metric as a function of the current and goal states and takes actions to reduce this error. In the context of grasping, the goal is typically encoded as a 6D pose to be achieved by the robot end effector.
A lot of progress has been made in the last decade in grasp learning to produce faster, more accurate and reliable grasp planners that output large number of grasp pose candidates. However, even when some of those grasp planners may be able to run in real-time, their outputs are not \textit{temporally consistent}, i.e., the output at any given frame is independent of the previous one, causing discontinuous jumps of the goal pose. The lack of temporal consistency makes it hard to design closed loop behaviors around these algorithm's outputs. It also poses a challenge for the downstream motion planner that consumes this goal pose. Moreover, in the case of eye-in-hand systems, the images obtained as the manipulator gets close to the object may fall outside of the training distribution of these planners, causing them to fail. 

\begin{figure}[t]
\centering
\includegraphics[width=0.42\textwidth]{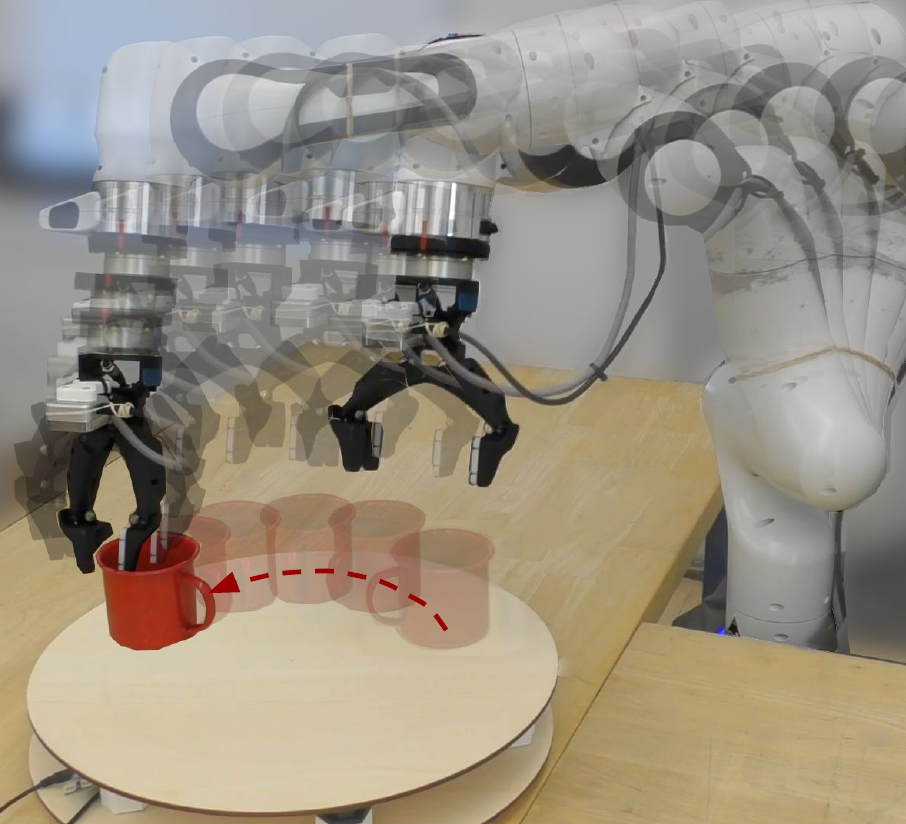}
\caption{\ourmethod{} allows servoing a robot manipulator to a successful grasp on a target object in a closed loop manner. It can execute grasps on slow moving objects and refine initial grasp proposals on static objects.}
\label{fig:main_figure}
\vspace{-5mm}
\end{figure}

In this paper, we address the problem of driving a robot manipulator to a successful grasp on a target object in a closed loop manner. A particular challenge for this task is to provide suitable scene information (for feedback) at all times during the manipulation task. If we only consider the use of a fixed external camera, the robot motion may produce occlusions as it navigates to execute the grasp. A wrist-mounted camera provides the best occlusion-free perspective at the moment of executing a grasp, but cannot possibly keep the target grasp in view at all times while the robot moves due to kinematics. Therefore, in this work, we limit ourselves to providing a closed loop mechanism to drive the gripper from a pre-grasp position to a final successful grasp using the visual feedback from a wrist-mounted camera (Figure \ref{fig:main_figure}).

Our approach begins with an initial grasp proposal provided by any available grasp planner with a global view of the scene as input. We allow the robot to navigate to a pre-grasp position, that is, a retracted pose from where if the gripper moves forward, it can execute the grasp. We leverage the fact that, in most cases, this grasp proposal will be either correct or close to correct. Therefore, if we search in a small neighborhood around this \textit{seed grasp}, our local grasp planner should be able to find a high quality grasp. Once we have found this grasp using the local information provided by the wrist camera, our task becomes that of ``rediscovering'' the same high quality grasp on the next frame.  

To search for the highest quality grasp in this region, we sample many grasp candidates around the \textit{seed grasp} and evaluate their quality. We use a Grasp Evaluator network trained with synthetic data which operates on the raw unsegmented point cloud data from the wrist camera. This network can provide a quality metric for each grasp and, combined with other scoring heuristics, allows us to produce a final score for each sampled candidate. The highest scoring grasp becomes the \textit{seed grasp} in the next iteration of the algorithm. 

We propose two mechanisms to improve the performance of our sampling-based approach. First, based on the previous seed grasps scores, we dynamically scale the sampling region size and the number of samples in it: we scale up when no good grasps are found, and scale down when a good grasp is found. Second, we make use of a Motion Vector Field Estimator and utilize the motion of points in the vicinity of the current grasp as a signal of object movement and use it to bias the center of our sampling region (the seed grasp) accordingly. 

The main contributions of our work are as follows:
\begin{itemize}
    \item {\it A sampling-based closed loop grasping algorithm}: Our algorithm takes in RGB-D inputs from a wrist camera and iteratively finds the highest quality grasp in a small region around a seed grasp. Unlike traditional grasp planners, our algorithm is designed from the ground up to output a temporally consistent high quality grasp. When initialized with an appropriate seed, the output of our algorithm can be consumed downstream by a suitable controller to drive a robotic gripper towards a successful grasp despite object disturbances.
    
    \item {\it Adaptive sampling}: We present two mechanisms that aid the efficiency of our sampling search mechanism. We linearly scale both the sampling region size and number of samples based on previous grasp scores to quickly recover when losing track of a grasp. We leverage a motion vector field estimator to bias the center of our sampling region to improve tracking of moving objects. 
    
    \item {\it Grasp scoring mechanism}: We show a simple scoring heuristic which takes into account the inherent grasp quality of a candidate but applies penalties to promote grasp temporal consistency. Additionally, we quantify and penalize the uncertainty in the grasp quality through the injection of synthetic noise into our Grasp Evaluator network. 
    
\end{itemize}

\ourmethod{} is the first to track grasps in a temporally consistent manner in 6 DoF at 20 Hz while also refining the grasp quality iteratively. We demonstrate that a system running our algorithm can improve grasping performance on both static and dynamic scenes. 

\section{Related Work}
In this section, we first review relevant work on grasp learning, followed by the progress in incorporating learning in closed-loop grasping with visual feedback.

\begin{figure*}[t]
\centering
\includegraphics[width=0.95\textwidth]{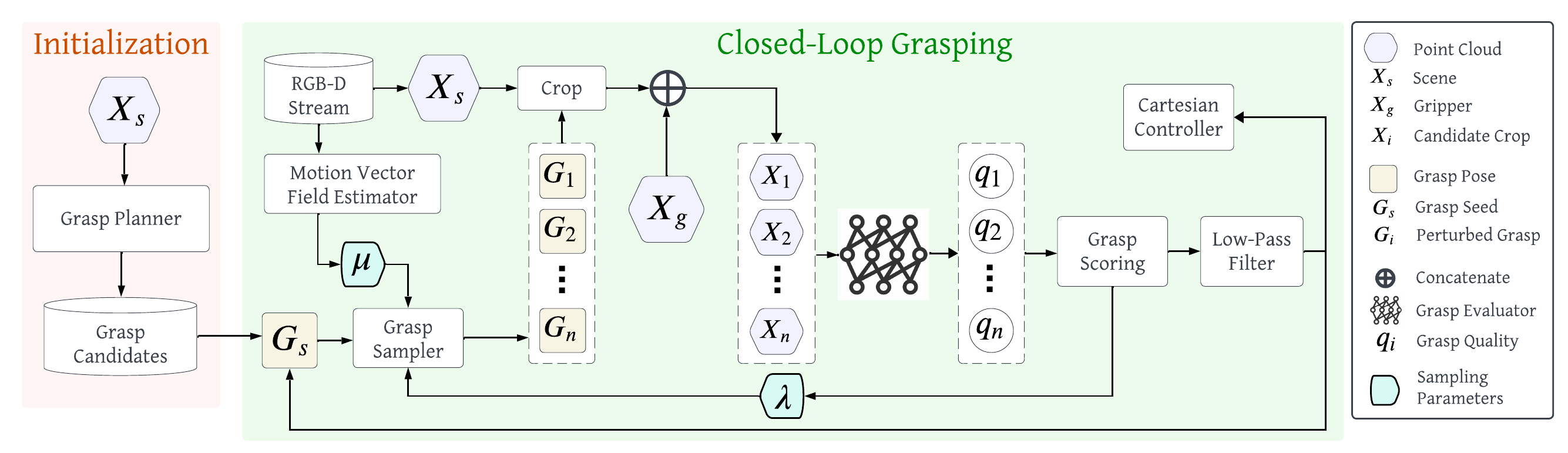}
\caption{A high level view of our method. We require an initial grasp candidate to serve as the grasp seed in our algorithm. Once the robot has navigated to a pre-grasp position for this initial proposal, we activate our control loop which updates the seed grasp continuously by finding the highest scoring grasp in a region around the seed.}
\label{fig:flow_chart}
\vspace{-3mm}
\end{figure*}

\subsection{Grasp Learning}
Classic research on robotic grasping treats the problem as a planning or optimization problem where certain geometric or mechanical constraints are considered for the grasping contact regions~\cite{murray2017mathematical, prattichizzo2016grasping}. Recent data-driven approaches explored grasp detection and evaluation without explicitly modeling these constraints. A lot of works in grasp learning focus on learning SE(2) grasps. For example, in~\cite{kumra2017robotic}, an input RGB image is mapped to a vector that encodes a good grasp in SE(2). Learning for SE(2) grasps has been demonstrated to be quite successful when equipped with high capacity visual encoders\cite{redmon2015real, johns2016deep}. More recently, grasp learning in SE(3) has gained popularity and is often tackled with more advanced network models. 
Gualtieri \etal \cite{gualtieri2016high} presented one of the earliest works on grasp learning in six dimensions. In their work, a binary classifier is adopted to evaluate the heuristics-based grasp proposals. Besides point cloud, many different volumetric representations have also been demonstrated for grasp learning, including voxel \cite{cai2022real}, signed distance function (SDF) \cite{breyer2021volumetric} and graphs \cite{lou2022learning}. Shape completion has also been explored to improve the quality of grasp detection \cite{varley2017shape, lundell2019robust}. In this work, we choose point cloud to represent the geometry as it not only preserves fine details of object shape information, but also has been well-explored for grasp learning in real-time.

Grasp learning on point cloud can be further categorized to two domains: grasp detection and grasp evaluation.
In \cite{qin2020s4g} and \cite{sundermeyer2021contact}, the authors applied PointNet++~\cite{qi2017pointnet++} to encode point features and infer grasps around the points. Zhao \etal \cite{zhao2021regnet} further extended the works by adding another "grasp region network" to infer grasp orientation as a categorical distribution for points determined to have a high grasp quality. On the other side, similar network structures have been proposed in \cite{mousavian2019}, \cite{murali20206} and \cite{yang2021reactive} to evaluate the quality of grasp proposals.
Grasp detection usually assumes no prior information about grasp candidates from previous time steps, therefore ensuring temporal consistency becomes a challenge. Since this work is not intended to solve end-to-end grasp tasks and we assume some priors to initialize the system, we focus on learning a good grasp evaluation function and rely on adaptive sampling to ensure spatial and temporal consistency.

\subsection{Closed-Loop Grasping}
Grasping in a closed-loop manner becomes important when the system has to deal with perception errors and object disturbances. Early works in this aspect attempt to guide the robot via visual servoing or object tracking, and the grasps are limited from top-dowm \cite{allen1993automated, houshangi1990control, smith1997grasping}. Some more recent works adopt similar ideas. For example, Marturi \etal \cite{marturi2019dynamic} proposed a work that explicitly tracks 6DoF object pose and combines this with grasp poses computed offline to achieve dynamic grasp planning. Other approaches such as \cite{menon2014motion}, \cite{rosenberger2020object} and \cite{kim2014catching} requires prior knowledge of the object shape thus are more restricted for real-world applications. Our closed-loop grasp system estimates motion at the scene level. Thus it does not require prior information of the object, neither does it explicitly track the object motion, making it more general in real-world scenarios.

In \cite{morrison2020learning}, the authors proposed generative grasping convolutional neural network (GG-CNN) for SE2 grasping and claimed that a lightweight network ruining at high frequency could enable closed-loop grasping scenarios. Our system follows similar principles by making the grasp evaluation module computation efficient and capable of running at real-time ($20Hz$). Therefore, instead of iteratively performing grasp detection across the whole scene and relying on similarity metrics to ensure temporal consistency \cite{fang2023anygrasp, liu2023target}, we choose to adaptively sample around the seed grasp from last frame and evaluate the grasps candidates in real time. From this perspective, our work is closely related to that of Yang \etal\cite{yang2021reactive}. Our approach differs in that we sample many candidates in both translation and rotation around a single seed (as opposed to a single perturbation in translation around many seeds). We also gather information from the scene-level motion vector field and use grasp scores history to dynamically adjust the sampling parameters. Lastly, our system is not limited to a single task and can be applied to other applications without further tuning. 
\section{Methodology}

We assume a grasp planner has selected a grasp candidate to be executed and the robot has successfully navigated to a pre-grasp location which is retracted from that target grasp $G_s$. Using the RGB-D information provided by a short range camera mounted on the wrist of the robot (RealSense D405), our high level objective is to servo the gripper from this pre-grasp location to a successful grasp on the target object in a closed loop manner.

 Provided the initial grasp candidate is in the vicinity of an actual high quality grasp, we posit that a grasp evaluator network will converge to this grasp or a similar one by continuously evaluating a set of randomly sampled grasps around the seed grasp $G_s$. More specifically, we aim to develop a real-time algorithm which takes in a seed grasp pose and a frame of RGB-D data from the wrist camera and outputs the highest quality grasp in a region around the seed grasp. Applying such algorithm frame by frame results in $G_s$ being constantly updated and a suitable controller can therefore drive the robot gripper to a successful grasp.
 

 In the next few sections we will provide more details about the full pipeline, which is shown in Figure \ref{fig:flow_chart})

\subsection{Grasp Sampling}
\label{sec:grasp_sampling}

Our pipeline begins with a \textit{seed grasp} $G_s$. During the first iteration, this seed grasp is provided by some off-the-shelf grasp planner. The assumption behind our sampling strategy is that this initial grasp proposal, while not perfect, is usually close to a high quality grasp. 

\begin{figure}[t]
\centering
\includegraphics[width=0.40\textwidth]{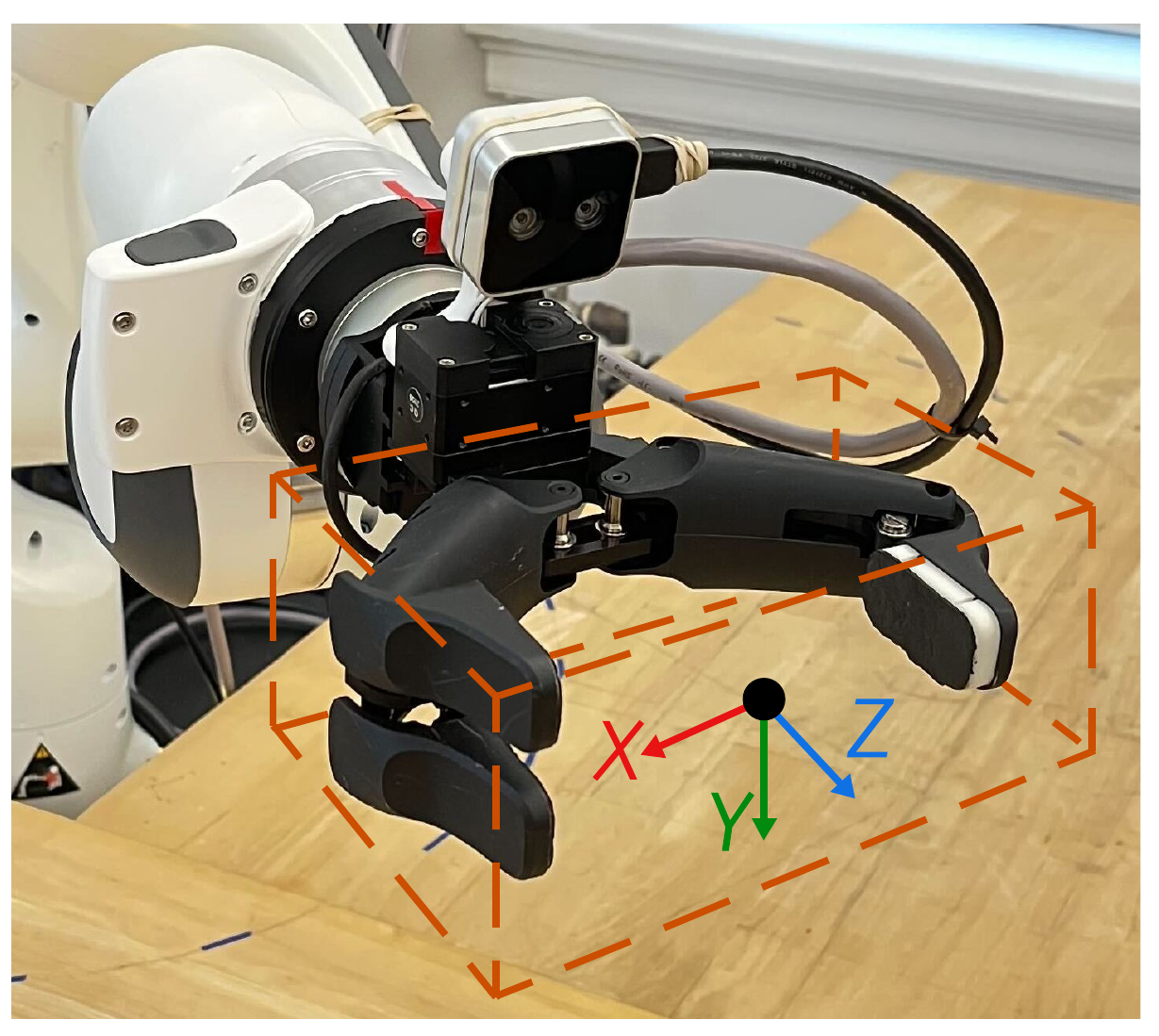}
\caption{Our robotic gripper along with the crop box used to segment the scene point cloud. The coordinate system shown is used to encode the concatenated scene and gripper point clouds which constitute the input to our Grasp Evaluator network.}
\label{fig:gripper_crop}
\vspace{-3mm}
\end{figure}

Given $G_s$, we would like to sample a set of $N$ grasps that lie in a small region around it. To do this, we generate $N$ transformation matrices where the translation vector is sampled from a uniform distribution with $\pm2~cm$ on each axis. In similar fashion, the rotation matrix is generated from a uniform distribution of Euler angles with zero degrees for roll and  $\pm5^{\circ}$ for both pitch and yaw. In our convention (see Figure \ref{fig:gripper_crop}, the roll axis matches the finger closure direction and therefore is not such an important factor in grasp quality. Our full set of perturbed grasps can then be obtained by right multiplying $G_s$ with these transformations matrices.

The sampling region limits mentioned above determine what we call our \textit{nominal sampling region}. However, we can dynamically scale $N$ and the region limits at run-time based on the circumstances. In practice, we scale both the region size and the number of perturbed grasps $N$ using a fixed scaling factor $\lambda$. If on a given iteration no good grasps are found as determined by our grasp scoring policy (all scores less than 0.5), we scale up the sampling region and $N$ by 30\% ($\lambda=1.3$) on the next iteration. This scaling may continue up to triple the nominal size. As soon as an iteration yields a good enough grasp (a score greater than 0.5), we revert the sampling region to its nominal size.

As described so far, our algorithm relies exclusively on the sampling region size to keep track of a moving object. Consider the simplified case of an object with only a single successful grasp pose moving at a constant speed. If we have discovered this high quality grasp at frame $t$, our only hope to recover this grasp at frame $t+1$ is that the sampling region is large enough to account for the object translation during that time interval. However, if we had a mechanism to keep track of the object movement, we can bias the pose around which we sample grasps and improve our tracking performance. This is the role of the motion field estimator.

We use GMFlow \cite{xu2022gmflow}, a transformer-based optical flow estimation algorithm on the RGB feed from our wrist camera. We lift the flow field from the image domain to our point cloud by taking the difference in depth values per pixel across two frames, creating a dense 3D vector field that describes the motion of the corresponding 3D points. At each iteration, we collect the 3D flow for all points in a sphere of radius $\rho$ around $G_s$, average them and obtain a single 3D velocity vector which, multiplied by the iteration time period produces the translational offset which we apply on the following iteration to $G_s$. 

The dynamic size of our sampling region in combination with the use of a motion vector field to bias the region center is what we refer to as \textit{adaptive sampling}.

\subsection{Grasp Evaluation Network}
\label{sec:grasp_evaluator}

Once we have a set of $N$ perturbed grasps, we require an evaluation network that can provide a quality metric for each. This grasp quality metric $Q \in [0,1]$ is one of the main factors used to determine the final score of a grasp (more details in section \ref{sec:grasp_scoring}).

We draw inspiration from GraspNet to design this Grasp Evaluator network. The backbone of the network is based on PointNet++, followed by a fully connected MLP with a sigmoid activation function applied at the output. Like GraspNet, the input to this network is the combination of the observed point cloud by the wrist camera and the gripper point cloud, each labeled using an extra binary feature indicating whether the point belongs to the scene (0) or to the gripper (1). The gripper point cloud is obtained by uniformly sampling the gripper proximal and distal link meshes. Unlike GraspNet, we decide to crop the scene pointcloud to a smaller region of interest around the grasping area. Using the grasp coordinate frame, we crop the scene pointcloud with a rectangular prism with limits $x = \pm 10~cm$, $y = \pm 5~cm$ and $-10<z<3~cm$ (see Figure \ref{fig:gripper_crop}).

To train our Grasp Evaluator, we use the ACRONYM dataset \cite{eppner2021acronym}. We utilize Isaac Gym to spawn objects and our gripper in the location corresponding to the grasp poses from ACRONYM. From each grasp position, we render the depth image from the wrist camera view, generate the corresponding point cloud and apply the same crop parameters explained above. These cropped point clouds are stored alongside the grasp label to build our training dataset.

All grasps from ACRONYM are generated on the object mesh surface. In our application, it is entirely possible to query grasp candidates (by our grasp sampler) which are more retracted from the object, in collision with it or even in an empty area with the object far away from the grasp. For this reason, we need to augment the data to capture these cases and label them accordingly. Following GraspNet convention, we call these \textit{hard negative} grasps. We generate them by applying both positive and negative offsets in the grasp Z axis (see Figure \ref{fig:gripper_crop}) as well as translations on the X,Y axes from the original set of grasps. Additionally, we verify that none of these generated grasps are similar to those in the original set (using Euclidean distance as similarity metric).

There are two data augmentations we perform to help bridge the sim2real gap. First, at training time, we add Gaussian noise to our point clouds with zero mean and 2mm of standard deviation on each axis. Additionally, our dataset contains the normal vectors for each point cloud. If the angle between the camera normal and a given point normal is between 80 and 90 degrees, we drop this point with a probability of 70\%. This is meant to represent the fact that most real depth cameras do not reliably provide depth information on surfaces which are at a shallow angle with respect to the camera normal direction. 

\subsection{Grasp Scoring}
\label{sec:grasp_scoring}

Our grasp evaluator network outputs a quality metric $Q \in [0,1]$, where 1 represents a high quality grasp and 0 a poor quality grasp. Because our network is trained as a classifier, the distribution of Q is heavily biased towards the values of 0 and 1. The underlying function our network approximates is complex and may have sharp discontinuities since small perturbations of the grasp pose can cause a grasp to go from success to failure. When evaluating grasps in such regions, even small amounts of noise in the point cloud can drive the quality towards either side of the classifier decision boundary, potentially creating a large spread in the quality metric. We consider that a true high quality grasp is that which can endure small perturbations while still producing a successful grasp and therefore we want to penalize grasp candidates which present large spreads in quality due to noise.

In such cases, a naive solution can be to take many measurements to average out the noise. However, this would hurt our real time performance. We propose instead to inject synthetic noise to the measured point cloud. For each grasp candidate $G_n$ and its corresponding point cloud $X_n$, we apply gaussian noise using the same parameters as during data augmentation at train time. We create $k$ copies of $X_n$, each with a different noise applied to it, and create a larger batch of inputs for network inference. After the forward pass through the evaluator network, we collect the mean quality $\bar{q_n}$ for each candidate, as well as the spread in quality scores $q_{sp} = max(Q_n)-min(Q_n)$ where $Q_n$ represents the set of $k$ quality values corresponding to grasp candidate $G_n$

We design a scoring function that penalizes large changes to $G_s$ to promote temporal consistency. To achieve this, we compute distance metrics for translation $T$ and rotation $R$ between the seed and the candidate grasp and multiply each with penalty terms $k_1$ and $k_2$ respectively. The translation distance metric $T$ is simply the $L2$ norm between the position of $G_s$ and $G_n$. For the rotation distance, we compute the rotation between $G_s$ and $G_n$ as $R_{sn} ={R_s}^T R_n$ where R represents the rotation matrix for each grasp. Then we compute the axis-angle representation as $Tr(R_{sn}) = 1 + 2 cos(\theta)$ and use the value of $\theta$ as the rotation distance $R$:

\vspace{-3mm}
\begin{equation}
S_n = \bar{q_n} - k_1*T - k_2*R - k_3*q_{sp} 
\end{equation}
\vspace{-4mm}

The score $S_n$ corresponding to a grasp candidate $G_n$ is then mainly composed of the mean quality and penalized by three different terms corresponding to translational and rotational distance from the seed and quality spread. The penalty weights $k_1$ through $k_3$ allows us to adjust each term's influence on the final score. 

The highest scoring grasp is fed to a simple low pass filter \cite{casiez_one_euro2012} to further enforce smoothness in the pose change of $G_s$ over time before updating the goal for the cartesian controller which servos the gripper.

\section{Evaluation and Results}

To evaluate our method, we design three experiments. First, we want to quantify how \ourmethod{} can improve grasping performance on a static scene. Second, we would like to evaluate our ability to track a moving object and then proceed to grasp it. Lastly, we attempt a human to robot handover task, where the object pose is completely unrestricted. 

In order to perform these experiments, we utilize a Franka Emika Panda robot arm equipped with a custom 4-bar linkage gripper. The robot is placed right next to a table where we place the objects for the static case or the rotating table in the moving objects case.

As explained previously, our method requires an initial grasp proposal for initialization. For both the static and moving objects experiments, we utilize Contact-GraspNet \cite{sundermeyer2021contact} (CGN) to provide the initial grasp seed $G_s$. 

Additionally, we need a fast controller capable of driving the gripper towards a moving target as our algorithm updates the best possible grasp at each iteration. For this task, we utilize a simple cartesian controller with a proportional gain to reduce the error between the current pose and the target pose. It must be noted that we can't naively servo the gripper directly towards the goal as we must approach the grasp pose from the pre-grasp position to avoid colliding with the object. Our grasping logic is such that we first target the pre-grasp position and if we can track it within certain tolerance for 5 iterations, we change the target to the final grasping pose. Note that in the current iteration of this work, this movement towards the final grasp pose is done in an open loop manner, due to the wrist camera inability to provide a reliable point cloud when the object geometry surpasses the gripper finger tips.

\subsection{Static objects}
\label{sec:static_grasping}

\begin{table*}[t]
\centering
\caption{Grasp Success rate - Static tabletop clearing task}
\label{static_table}
\begin{tabular}{lcccccccccc}
\hline
\textbf{Grasp planner} & \multicolumn{1}{l}{\textbf{Bowl}} & \multicolumn{1}{c}{\textbf{CheezIt}} & \multicolumn{1}{c}{\textbf{Bottle}} & \multicolumn{1}{c}{\textbf{Can}} & \multicolumn{1}{c}{\textbf{Pringles}} & \multicolumn{1}{c}{\textbf{Mug}} & \multicolumn{1}{c}{\textbf{Mustard}} & \multicolumn{1}{c}{\textbf{Small box}} & \multicolumn{1}{c}{\textbf{Total}} \\  \hline 
\\[-3mm] \hline
\\[-2mm]

CGN         & 15/15   & 10/15   & 7/15   & 4/15   & 11/15   & 8/15    & 9/15    &  8/15    & 60\%   \\
CGN+VFAS    & 15/15   & 15/15   & 8/15   & 14/15  & 12/15   & 12/15   & 11/15   &  14/15   & 84\%

\end{tabular}
\end{table*}

\begin{figure}[t]
\centering
\includegraphics[width=0.40\textwidth]{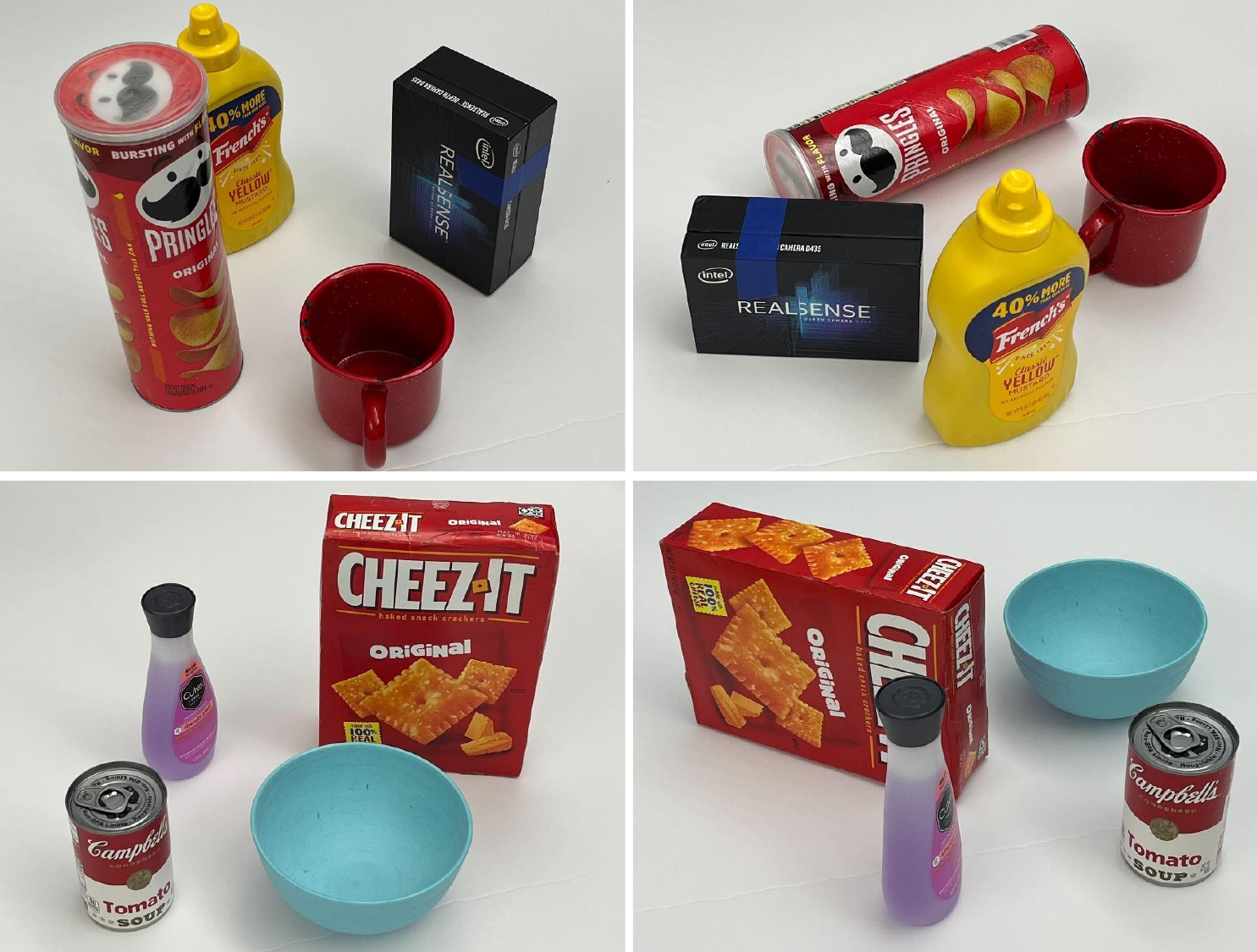}
\caption{The object set used in our experiments. We show here four example scenarios used in our static object test. Subsets of these objects are used as well for the moving object and human to robot handover tests}
\label{fig:objects}
\end{figure}

For this experiment, the goal is to clear all the objects on the table. We utilize a total of 8 objects, with a maximum of 4 on the table for a given scenario. We define 5 scenarios with a subset of 4 objects and another 5 with the remaining 4 objects for a total of 10 scenarios (Figure~\ref{fig:objects}). Each scenario presents a different arrangement of the objects on the table. We record the grasp success rate for each object across the 10 scenarios, where each scenario is attempted 3 times.

We first observe the scene from a position where all the objects are visible. We utilize a pre-trained and fine-tuned model of Mask R-CNN \cite{he2017mask} to segment the objects point cloud and query CGN for grasps on each object. We ranked the returned grasps based on their kinematic feasibility and score higher those which can be achieved with the robot arm further away from joint limits (since the cartesian controller cannot gracefully handle joint limits).

For our baseline, we record the grasp success rate that results from executing the grasps proposed by CGN without our algorithm. Then we repeat the experiment with CGN proposing a grasp, but using \ourmethod{} after the robot reaches the pre-grasp position to drive the gripper towards a refined grasp proposal. Results are shown in Table~\ref{static_table}. 

\ourmethod{} provides a significant improvement over the baseline across all objects, but especially on objects where small perturbations can result in failed grasp, such as the can which is typically grasped from the top.

\subsection{Moving objects}
\label{sec:dynamic_grasping}

\begin{table}[]
\centering
\caption{Grasp success rate - Moving object on a turntable}
\label{dynamic_object_results}
\begin{tabular}{lccccc}
\hline
\textbf{Object velocity} & \textbf{Mug} & \textbf{Can} & \textbf{Bowl} & \textbf{Bottle} & \textbf{Total} \\  \hline 
\\[-3mm] \hline
\\[-2mm]

\multicolumn{1}{c}{} & \multicolumn{5}{c}{\textit{Adaptive Sampling Disabled}} \\ \hline
\\[-2mm]

Low                      & 5/5          & 5/5          & 5/5           & 4/5             & \textbf{95\%}  \\
Medium                      & 4/5          & 2/5          & 4/5           & 2/5             & \textbf{60\%}  \\
High                     & 0/5          & 1/5          & 5/5           & 1/5             & \textbf{35\%} \\
\\[-2mm]

\multicolumn{1}{c}{} & \multicolumn{5}{c}{\textit{Adaptive Sampling Enabled}} \\ 
\hline
 \\[-2mm] 

Low                      & 5/5          & 3/5          & 5/5           & 5/5             & \textbf{90\%}  \\
Medium                      & 5/5          & 5/5          & 5/5           & 2/5             & \textbf{85\%}  \\
High                     & 5/5          & 3/5          & 5/5           & 3/5             & \textbf{80\%} 

\end{tabular}
\end{table}

In this experiment, we use a custom built turntable (Figure \ref{fig:main_figure}) to analyze the effect of our adaptive sampling strategies when tracking objects moving at different speeds. We place an object in a starting position at three different radial distances (6, 10 and 14cm) and use a fixed angular velocity of $\frac{2\pi}{10} rad/s$ to generate trials at a low, medium and higher speed (3.8, 6.3 and 8.8 cm/s respectively).

The robot will observe the scene and query CGN for grasps and move to the pre grasp position. At this point, the turntable is commanded to move in a random direction, with a goal position between $\pm60^{\circ}$ and $\pm120^{\circ}$. The robot needs to track the object throughout this movement and grasp it after it comes to a stop. The objects used in this experiment are the mug, bowl, bottle and can. We compare our algorithm grasping performance with and without adaptive sampling. When running without adaptive sampling, the motion vector field estimator and the dynamic sampling region resizing are disabled. 

Results in Table~\ref{dynamic_object_results} show that, as the object movement speed increases, our adaptive sampling strategies are fundamental to the final performance of the system. 

\subsection{Human to Robot Handover}
\label{sec:handover}

For our final experiment, we demonstrate that \ourmethod{} can also be used to enable human to robot handover tasks, where the object pose is completely unrestricted and may change in both translation and rotation. 

Unlike the previous experiments, we do not provide an initial grasp proposal to start tracking the object. Instead, we place the robot arm in a fixed position, hard-code the seed grasp to be 15cm in front of the gripper and let our algorithm run continuously searching for a good grasp in this area. During this initialization stage, $G_s$ is not updated. Once a good grasp is found (score greater than 0.5), we transition to the tracking phase where the seed is updated on each iteration to track the high quality grasp found in the previous iteration. 

The participants were instructed to pick any of 4 objects (bowl, bottle, small box, mustard) and slowly present the object to the robot in the area in front of the gripper. They may freely move and change the object pose as desired during the tracking phase. We consider the handover to be successful if the robot grasps the object within the first 20 seconds of starting the tracking phase. 

A total of 8 participants performed 3 trials with each object for a total of 96 trials (24 trials per object). The overall success rate was 81.25\% with 3, 5, 4 and 7 failures for the bowl, bottle, small box and mustard respectively.

\section{Conclusions}

We presented a new closed-loop grasping method \ourmethod. We demonstrated that it can effectively refine and track an initial grasp proposal solely using the feedback provided by an RGB-D camera mounted on the wrist of a robot manipulator. This is enabled by our real-time sampling strategy, which is capable of evaluating a large number of candidate grasps and scoring them to promote temporal consistency of the output target grasp. Results show that \ourmethod{} significantly improves the grasping performance for static objects and enables the possibility of grasping moving objects.

However, there are some limitations to our method. In its current form we cannot guarantee that the final grasp executed on the object will maintain the original grasp affordance. It is entirely possible that our algorithm will shift and drift the grasp on the object either moving within a continuous grasp manifold or even jump to a different one (for example with a mug, the grasp might ``walk'' through the rim and, in some cases, jump to the handle) based on how the object is moving. Additionally, due to the fact that no semantic information is provided, if objects are cluttered or moving in close proximity, the grasp proposal could jump from one object to another. Segmentation algorithms are, as of today, too computationally expensive and cannot be incorporated into our approach without a dramatic drop in our sampling frequency. Lastly, grasping faster moving objects likely will require an additional estimator which predicts the future movement of the target to provide a feed-forward mechanism in our control loop.

Perhaps one the biggest challenges of building a closed loop grasping system is the motion planner which needs to consume a constantly evolving goal pose. In our experiments, we use a cartesian controller to drive the robot movement due to its computational efficiency. However, this approach is completely unaware of the scene or the robot kinematics, which can cause the robot to collide with other objects or itself, as well as encountering joint limits during motion. A robust system will require a collision-aware motion planner which can replan in the order of tens of milliseconds while keeping the target object in view. 

\bibliographystyle{IEEEtran}
\bibliography{main.bbl}

\end{document}